\documentclass{article}

\usepackage{PRIMEarxiv}

\usepackage[utf8]{inputenc} 
\usepackage[T1]{fontenc}    
\usepackage{hyperref}       
\usepackage{url}            
\usepackage{booktabs}       
\usepackage{amsfonts}       
\usepackage{nicefrac}       
\usepackage{microtype}      
\usepackage{lipsum}
\usepackage{fancyhdr}       
\usepackage{graphicx}       
\graphicspath{{media/}}     

\title{Forge-and-Quench: Enhancing Image Generation for Higher Fidelity in Unified Multimodal Models}
\author{Yanbing Zeng\thanks{Equal contribution.}\quad Jia Wang\footnotemark[1]\quad Hanghang Ma\quad Junqiang Wu\quad Jie Zhu\quad Xiaoming Wei\quad Jie Hu\thanks{Corresponding author and project leader.} \\
Meituan\\
\texttt{\{hujie39\}@meituan.com} \\
}

\begin{document}
\maketitle

\begin{abstract}
Integrating image generation and understanding into a single framework has become a pivotal goal in the multimodal domain. However, how understanding can effectively assist generation has not been fully explored. Unlike previous works that focus on leveraging reasoning abilities and world knowledge from understanding models, this paper introduces a novel perspective: leveraging understanding to enhance the fidelity and detail richness of generated images. To this end, we propose \textbf{Forge-and-Quench}, a new unified framework that puts this principle into practice.
In the generation process of our framework, an MLLM first reasons over the entire conversational context, including text instructions, to produce an enhanced text instruction. This refined instruction is then mapped to a virtual visual representation, termed the \textbf{Bridge Feature}, via a novel \textbf{Bridge Adapter}. This feature acts as a crucial link, \textit{forging} insights from the understanding model to \textit{quench} and refine the generation process. It is subsequently injected into the T2I backbone as a visual guidance signal, alongside the enhanced text instruction that replaces the original input.
To validate this paradigm, we conduct comprehensive studies on the design of the Bridge Feature and Bridge Adapter.
Our framework demonstrates exceptional extensibility and flexibility, enabling efficient migration across different MLLM and T2I models with significant savings in training overhead, all without compromising the MLLM's inherent multimodal understanding capabilities. Experiments show that Forge-and-Quench significantly improves image fidelity and detail across multiple models, while also maintaining instruction-following accuracy and enhancing world knowledge application. Models and codes are available at \url{https://github.com/YanbingZeng/Forge-and-Quench}.
\end{abstract}

\section{Introduction}
While models for image generation and understanding have achieved remarkable capabilities, recent research has increasingly focused on their unification within a single, cohesive framework. Some approaches~\citep{sun2023emu, sun2024generative, team2024chameleon, fan2025unified, wu2024vila} concentrate on tokenizing data from different modalities for a unified autoregressive model. While the form is concise, such an approach requires a significant training cost. Alternatively, another paradigm~\citep{ge2307planting, pan2025transfer, chen2025blip3} freezes the pre-trained Multimodal Large Language Model (MLLM) and text-to-image (T2I) models, and connects them using a lightweight adapter, which is trained using significantly fewer computing resources. Beyond mere structural unification, a critical question is how these two capabilities can mutually enhance one another. Models like MetaQuery~\citep{pan2025transfer} and BLIP3-o~\citep{chen2025blip3} have shown success by linking an MLLM to a T2I model, effectively transferring the MLLM's reasoning and world knowledge to the generation process. Benefiting from these, this paradigm improves the generation quality without degrading the model's inherent understanding capabilities.

Despite this success, we recognize that the current paradigm is an early step in how understanding can facilitate generation. The prevailing approach treats the MLLM as a sophisticated prompt rewriter, implicitly enhancing the initial text instruction before handing it off to a fixed denoising process. This one-time `handoff' mechanism, however, can create an informational bottleneck. It forces the MLLM to compress a wealth of multi-faceted visual knowledge, such as nuances in texture, lighting, and composition, into a single semantic embedding, where fine-grained details may be lost or entangled.
This observation motivates us to move beyond using the MLLM as a mere ``prompt rewriter''. In response, we propose a deeper integration where the MLLM actively participates in and guides the generative process.

Our design is inspired by advancements in controllable generation~\citep{zhang2023adding, ye2023ip, lian2023llmgrounded}, particularly methods like IP-Adapter~\citep{ye2023ip}. While IP-Adapter is designed to preserve the identity of a reference image, we observe a crucial side effect: its visual guidance significantly enhances the fidelity and detail of the generated output. This powerful mechanism, however, is unavailable in standard T2I tasks which lack a reference image.
This gap leads to our core hypothesis: \textbf{can an MLLM learn to \textit{forge} a virtual visual signal from the text instruction alone?} Our central thesis is that such a signal, when injected through a lightweight adapter, can replicate the fidelity boost of image-based conditioning without requiring an actual reference image.

To overcome this limitation, we introduce Forge-and-Quench, a framework that redefines the synergy between understanding and generation. Moving beyond the single handoff paradigm, our method tasks an MLLM with \textit{forging} two complementary signals: a semantically-rich text instruction and a powerful virtual visual feature, termed \textbf{Bridge Feature}, through \textbf{Bridge Adapter}. While the enhanced text provides high-level guidance informed by the MLLM's reasoning, the Bridge Feature is subsequently injected into the T2I backbone via a \textbf{Injection Adapter} to steer the synthesis with fine-grained visual details. This dual-conditioning strategy is engineered to deliver substantial gains in image fidelity and detail richness. Our main contributions are:
\begin{itemize}

\item We propose Forge-and-Quench, a unified architecture that enriches refined text conditioning with extracted MLLM visual signals via a dual-path design. This approach significantly enhances image fidelity while fully preserving the MLLM's understanding capabilities.
\item We conduct a rigorous analysis of the Bridge Adapter and Bridge Feature, establishing design principles for effectively leveraging understanding to enhance generation.

\item We propose a lightweight Injection Adapter that integrates our Bridge Feature into diverse T2I backbones, regardless of their text encoders. This ensures seamless extensibility and broad applicability with minimal training overhead.
\end{itemize}

\section{Related works}
\textbf{Unified multimodal models.} 
Recent unified multimodal models have explored diverse strategies for integration. The ambitious ``Any-to-Any" paradigm, pioneered by NExT-GPT~\citep{wu2024next}, introduces an LLM-centric architecture where various modalities are projected into a frozen LLM, which is then fine-tuned via a lightweight LoRA adapter. To deepen the interaction between modalities, some works like Emu~\citep{sun2023emu}, Emu2~\citep{sun2024generative}, Chameleon~\citep{team2024chameleon}, UniFluid~\citep{fan2025unified}, and VILA-U~\citep{wu2024vila} aim to unify representations at the token level with an autoregressive objective, focusing on building large, deeply trained models. Another approach involves complex fused architectures that combine different mechanisms, such as diffusion/flow models for images with autoregressive parts for text, within a single integrated framework, as seen in Transfusion~\citep{zhou2024transfusion}, Mogao~\citep{liao2025mogao}, and Bagel~\citep{deng2025emerging}. This architectural complexity is further exemplified by other intricate designs, such as the dual-encoder architecture of Janus/Janus-Pro~\citep{wu2025janus, chen2025janus}. While powerful, these models typically require extensive and costly joint pre-training.

An alternative, some approaches avoid this cost by bridging powerful, pre-trained models. SEED series~\citep{ge2307planting, ge2023making, ge2024seed} involve fine-tuning a Large Language Model (LLM) to predict discrete visual tokens from a specialized image tokenizer, thereby deeply integrating the LLM into the visual planning process. Furthermore, a more lightweight and efficient strategy, used in MetaQuery~\citep{pan2025transfer} and BLIP3-o~\citep{chen2025blip3}, keeps both the MLLM and the diffusion model fully frozen. They train only an adapter to extract and transfer continuous embeddings for generation. However, they focus mainly on leveraging the MLLM's reasoning ability and world knowledge, and inject the MLLM embeddings through existing T2I pathways. As a result, the guidance from the MLLM primarily impacts macro-level features, such as object composition and spatial arrangement, while failing to improve finer details.

\textbf{Diffusion-based T2I generation.}
Diffusion-based T2I models achieve a significant leap in generation quality by establishing a core paradigm: conditioning on powerful, pre-trained language backbones~\citep{nichol2021glide, ramesh2022hierarchical, rombach2022high}. Imagen~\citep{saharia2022photorealistic} shows that a stronger text encoder often contributes more to fidelity than a larger diffusion model. This paradigm is advanced by models like SDXL~\citep{podell2023sdxl}, which use a dual text-encoder for superior prompt comprehension, and other works that push aesthetic boundaries~\citep{kolors, midjourney-v6.1, cai2025hidream}. More recent works focus on novel architectures for deeper integration, like the MMDiT architecture in Stable Diffusion 3~\citep{esser2024scaling} and FLUX.1~\citep{blackforestlabs_flux}. Subsequent works~\citep{gao2025seedream, wu2025qwen} further enhance multiple stages of the generation process, leading to significant improvements in overall image quality. Given that these powerful backbone models are developed at a significant cost and exhibit exceptional generative capabilities, how to best leverage them has become a key point.

\textbf{Information injection for controllable generation and editing.}
A complementary line of research addresses controllable image generation and editing by injecting auxiliary conditions into diffusion models~\citep{wang2025image}. ControlNet~\citep{zhang2023adding} introduces external structural controls (e.g., edges, depth), enabling coarse-grained and spatially precise generation. For finer control, IP-Adapter~\citep{ye2023ip} uses reference images to preserve object identity, while T2I-Adapter~\citep{mou2024t2i} and the more generalized Composer~\citep{huang2023composer} enable compositional multi-attribute control by combining various conditions like style and layout. A growing trend involves leveraging MLLMs to process complex instructions. LLM-grounded Diffusion (LMD)~\citep{lian2023llmgrounded}, for example, uses an LLM to parse prompts into a structured layout to improve spatial accuracy, while MGIE~\citep{fu2023guiding} employs MLLMs to enrich editing instructions. FreeEdit~\citep{he2024freeedit} further injects fine-grained reference features in a mask-free manner for high fidelity. Together, these works validate the potential of MLLMs for efficiently processing conditional information and enabling fine-grained detail injection.

\begin{figure*}[t]
\centering
\includegraphics[width=1.0\textwidth, trim=0cm 3.5cm 1.9cm 0cm, clip]{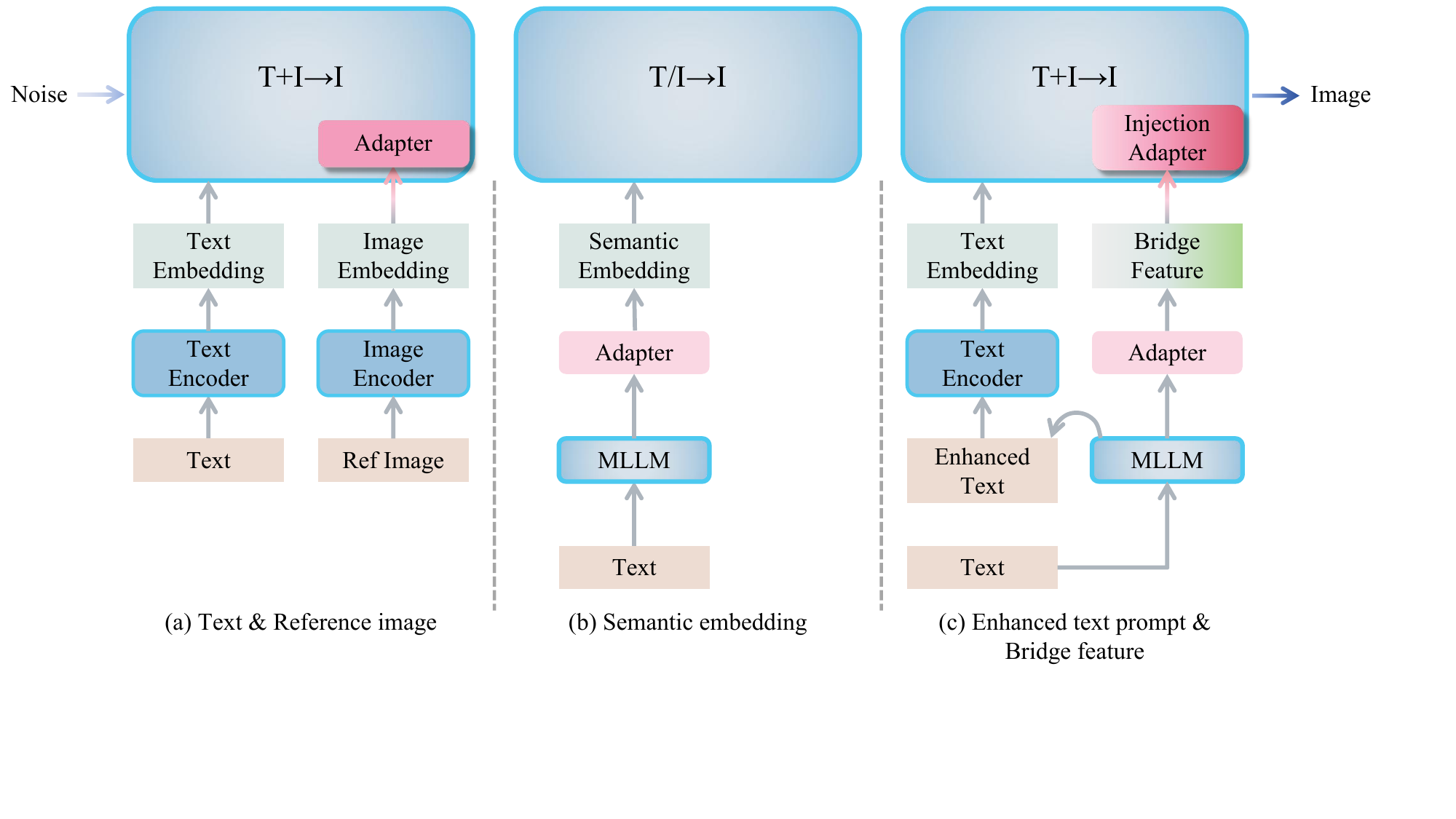} 
\setlength{\abovecaptionskip}{-5pt}
\caption{Three methods of image generation. (a) Given a text prompt and a reference image. (b) Given text, mapped to text/image semantic embedding using MLLM. (c) Given text, enhanced text and image semantic embedding (Bridge Feature) obtained using MLLM.}
\label{fig:core_design}
\vspace{-10pt}
\end{figure*}

\section{Methods}

\subsection{Preliminary and Motivation}
To precisely state our framework, we first formalize the \textit{conditioning mechanisms} of three prominent paradigms in image generation. All these approaches share a common latent flow matching backbone, where a velocity prediction network $u_\theta$ in flow matching~\citep{lipman2022flow, liu2022flow} operates on a latent variable $z_s$ at timestep $s$. The key distinction lies in the conditional information used to guide the generation process.

\textbf{T2I generation and controllable image generation.}
Given a text prompt $t$, one or more text encoders $\mathcal{E}_T$ convert it into a conditional embedding $e_t = \mathcal{E}_T(t)$. The $u_\theta$ is then conditioned solely on this text embedding:
\begin{equation}
\hat{v}_{s-1} = u_\theta(z_s, s \,|\, e_t),
\end{equation}
where $\hat{v}_{s-1}$ denotes the predicted velocity at timestep $s$.

To enable finer control over the generation process, a reference image $i_r$ is introduced, which is then encoded into an image embedding $e_i = \mathcal{E}_I(i_r)$ via an image encoder $\mathcal{E}_I$. The $u_\theta$ is then jointly conditioned on both $e_t$ and $e_i$:
\begin{equation}
\label{eqn:image_condition}
\hat{v}_{s-1} = u_\theta(z_s, s \,|\, e_t, e_i)
\end{equation}
Our experiments confirm that when a real reference image is additionally provided, this dual conditioning scheme significantly enhances the fidelity and detail richness of images.

\textbf{Unified Models with Modal Bridge.}
Recent works, such as MetaQuery and BLIP3-o, adopt a different strategy by freezing both the MLLM and the T2I backbone. Given the text prompt $t$, these methods first extract an intermediate embedding $e_m = \mathcal{F}_M(t)$ from the MLLM using a set of learnable queries. This intermediate embedding is then mapped to a new semantic space by a modal bridge $\mathcal{B}$, resulting in a bridge embedding $e_b = \mathcal{B}(e_m)$. The $e_b$ is believed to encapsulate the MLLM's reasoning ability and world knowledge, which is subsequently used as the \textit{sole condition}, effectively replacing the original text embedding:
\begin{equation}
\label{eqn:mllm_condition}
\hat{v}_{s-1} = u_\theta(z_s, s \,|\, e_b).
\end{equation}

\textbf{Motivation of our work.}
Contrasting these paradigms raises an important question: while conditioning on an MLLM-derived embedding (Eq.~\ref{eqn:mllm_condition}) shows promise, it is unclear if this strategy fully leverages the potential of multimodal conditioning. We have observe that explicit image features (Eq.~\ref{eqn:image_condition}) can greatly enhance the realism and detail of generated images. However, in text-only generation scenarios, a real reference image is typically unavailable.

To bridge this gap, we propose a framework that empowers an MLLM to \textit{forge} a high-quality, virtual visual feature $e_b$ in the absence of a real reference image, and uses it to \textit{enhance} the T2I generation process rather than solely \textit{replace} the text condition. Our objective is to develop a conditioning scheme that integrates the complementary advantages of Eq.~\ref{eqn:image_condition} and Eq.~\ref{eqn:mllm_condition}:
\begin{equation}
\hat{v}_{s-1} = u_\theta(z_s, s \,|\, e_t^{*}, e_b)
\end{equation}
where $e_t^{*}$ is derived from the enhanced text prompt. 

\begin{figure*}[t]
\centering
\includegraphics[width=1.0\textwidth, trim=0.5cm 2.5cm 0cm 2.5cm, clip]{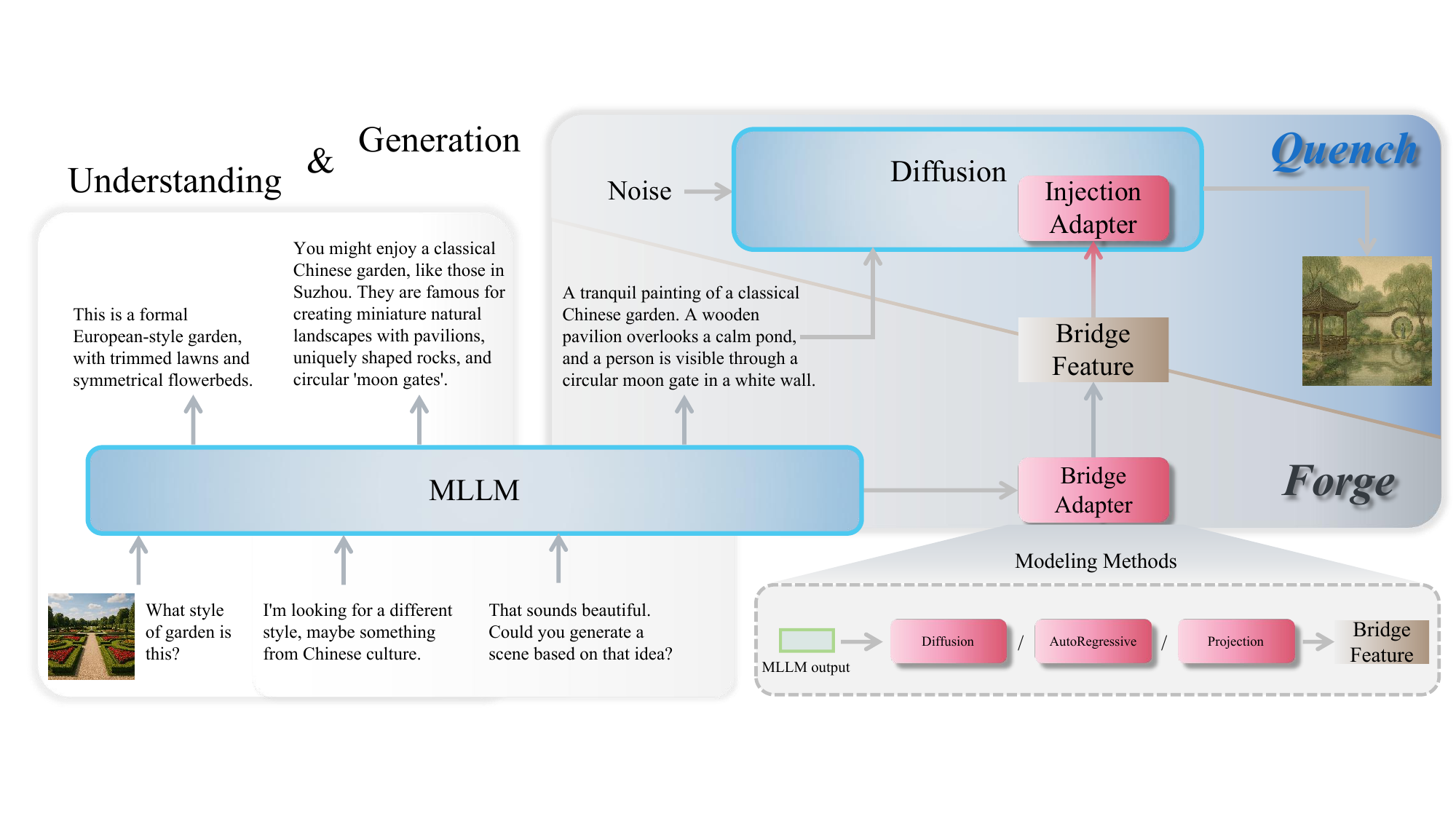} 
\caption{Forge-and-Quench, our unified framework.}
\label{fig:training_strategy}
\vspace{-10pt}
\end{figure*}

\subsection{Architecture Design}
The overall architecture of our framework, illustrated in Fig.~\ref{fig:training_strategy}, implements the ``(Enhanced Text + Virtual Image) $\rightarrow$ Image" generation paradigm. By freezing the MLLM, we ensure that its understanding capabilities are fully retained throughout the process. 
The generation pipeline consists of two parts. First, the MLLM is used to produce both an enhanced text prompt $t^*$ and $e_b$, thereby enriching the conditional information for image synthesis. 
Second, the $t^*$ are sent to T2I backbone with $e_b$ injected using a Injection Adapter. This modular design enables flexible integration and independent optimization of each stage.


\subsubsection{Forge}

\textbf{Text instruction enhancement.}
As shown in Fig.~\ref{fig:training_strategy}, our framework leverages the MLLM's advanced comprehension and world knowledge to semantically enrich $t$. Rather than simple paraphrasing, the MLLM is capable of understanding nuanced user intent, incorporating cultural context, and adapting to multi-turn interactions. 

For example, when a user expresses interest in a different garden style and references Chinese culture, the MLLM can suggest a ``classical Chinese garden" and further elaborate with culturally specific elements such as pavilions, moon gates, and rock arrangements.
Given the evolving conversation, the MLLM transforms a basic prompt like ``a formal European-style garden" into a much richer and contextually appropriate description, such as:  
``A tranquil painting of a classical Chinese garden. A wooden pavilion overlooks a calm pond, and a person is visible through a circular moon gate in a white wall." 


\textbf{Forging Bridge Feature.} 
We choose the SigLIP vision encoder~\citep{zhai2023sigmoid}, $\mathcal{E}_{\text{Sig}}$, to acquire $e_b$ for its strong visual representation capabilities.
Given a ground-truth image $I$, its target feature is defined as $e_s = \mathcal{E}_{\text{Sig}}(I)$. To forge this feature from text, we feed $t^*$ into the frozen MLLM and use a set of learnable queries, $\mathcal{F}_M$, to extract a fixed-length intermediate embedding $e_m = \mathcal{F}_M(t^*)$.

We then train a Bridge Adapter, $\mathcal{B}_\phi$ to learn a mapping from the MLLM's abstract embedding $e_m$ to $e_s$. The adapter learns to predict the flow $v_b=\epsilon-e_s$ added to a ground-truth SigLIP feature $e_s$ at a given timestep $k$, using the MLLM's output $e_m$ as the guiding condition. The training objective is formulated as:
\begin{equation}
\mathcal{L}_{\text{F}} = \mathbb{E}_{e_s, e_m, \epsilon \sim \mathcal{N}(0, I), k} \left[ \left\| v_b - \mathcal{B}_\phi(e_{s,k}, k, e_m) \right\|_2^2 \right],
\end{equation}
where $e_{s,k}$ is the noisy version of the target feature $e_s$ at diffusion step $k$. Once trained, the adapter can generate a high-quality Bridge Feature $e_b$ from any intermediate embedding $e_m$ via the reverse diffusion process. This allows us to effectively \textit{forge} a detailed visual feature directly from textual information.

\subsubsection{Quench}
In this step, both $t^*$ and $e_b$ are injected into the T2I model to guide the final image synthesis. We \textbf{freeze the entire T2I backbone} $u_\theta$, and train \textbf{only the lightweight Injection Adapter} $\mathcal{A}_\psi$. 
Specifically, $t^*$ is processed by the T2I model's native text encoder $\mathcal{E}_T$ to produce the standard text embedding $e_t^* = \mathcal{E}_T(t^*)$. And $e_b$ is passed through $\mathcal{A}_\psi$, and then injected to each DiT layer of $u_\theta$, typically through cross-attention mechanisms similar to IP-Adapter.

The training objective is to optimize $\mathcal{A}_\psi$ by predicting the flow $v=\epsilon-z_0$, which is now conditioned on both the text and the adapted visual feature:
\begin{equation}
\mathcal{L}_{\text{Q}} = \mathbb{E}_{z_0, t^*, e_b, \epsilon \sim \mathcal{N}(0, I), s} \left[ \left\| v - u_\theta(z_s, s \mid e_t^*, \mathcal{A}_\psi(e_b)) \right\|_2^2 \right],
\end{equation}
where $z_0 = \mathcal{E}_{\text{VAE}}(I)$ is the latent representation of $I$, and $\mathcal{E}_{\text{VAE}}$ is the VAE encoder. As $u_\theta$ is frozen, all gradients from this loss are used to update the weights of $\mathcal{A}_\psi$ exclusively. This process ensures that image generation is guided by both the precise semantics embedding $e_t^*$ and the rich visual priors of $e_b$, resulting in images with higher fidelity and richer detail.

\textbf{Inference pipeline:} 

1) \textbf{Forge:} given a user prompt $p$: The MLLM first enriches the initial prompt: $t \rightarrow t^*$. Then, use MLLM and $\mathcal{B}_\phi$ to generate $e_b$: $t^* \rightarrow e_m \rightarrow e_b$.

2) \textbf{Quench:} the T2I model, guided by the Injection Adapter $\mathcal{A}_\psi$, synthesizes the final image conditioned on both $e_t^*$ and $e_b$.

This modular design, where the core MLLM and T2I models remain frozen, allows for main components to be easily swapped. Flexibility and scalability are maintained by only needing to retrain the corresponding lightweight adapters.



\begin{table}
    \renewcommand{\arraystretch}{1.2}
    \caption{Automatic evaluation results.}
    \setlength{\abovecaptionskip}{-5pt}
    \label{tab:auto_metrics}
    \centering
    \resizebox{1.0\textwidth}{!}{ 
    \begin{tabular}{lccccc}
        \toprule 
             Method & COCO-30K FID $\downarrow$ &GPT-Fidelity $\uparrow$ & GenEval $\uparrow$ & DPG-Bench $\uparrow$ &WISE Score $\uparrow$ \\
        \midrule 
            MeiGen-Image &23.97 & 12\% win &0.7845 &85.94 & 0.55 \\ 
            MeiGen-Image-FaQ &19.86 & 88\% win &0.7837 &86.83 & 0.70 \\
        \addlinespace[3pt]
        \cline{1-6}
        \addlinespace[3pt]
            FLUX.1-dev &27.71 & 22\% win & 0.6518 &83.66 & 0.56\\ 
            FLUX.1-dev-FaQ &20.83 & 78\% win &0.6436 &83.01 & 0.66 \\ 
        \bottomrule
    \end{tabular}
    }
\end{table}

\begin{figure*}[t]
\centering
\includegraphics[width=1.0\textwidth, trim=0cm 0cm 0cm 0cm, clip]{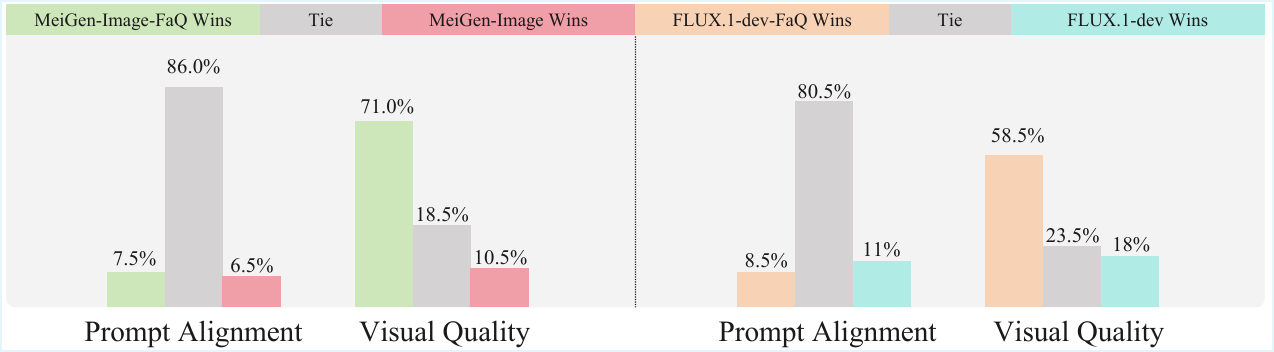} 
\setlength{\abovecaptionskip}{-0pt}
\caption{Human evaluation results.}
\label{fig:human_anno}
\vspace{-8pt}
\end{figure*}

\section{Experiments}

\subsection{Setup}
We validate our framework on two diverse T2I backbones: \textbf{FLUX.1-dev} and \textbf{MeiGen-Image}. The latter is a 6B-parameter internally model adopting a single-stream block and double-stream block architecture, slated for future release. Models enhanced by our method are referred to as `[Model-Name]-FaQ'.

Our framework's components are trained as follows. The 2B-parameter Bridge Adapter is trained for 500k steps on 200M image-text pairs. The 1B-parameter Injection Adapter is then trained for 80k steps on a 13M-sample subset, beginning at 512px resolution before concluding at 1024px. Full hyperparameters are detailed in Appendix~\ref{ap-sec:training_setting}.


\begin{figure*}[t]
\centering
\includegraphics[width=1.0\textwidth, trim=0cm 0cm 0cm 0cm, clip]{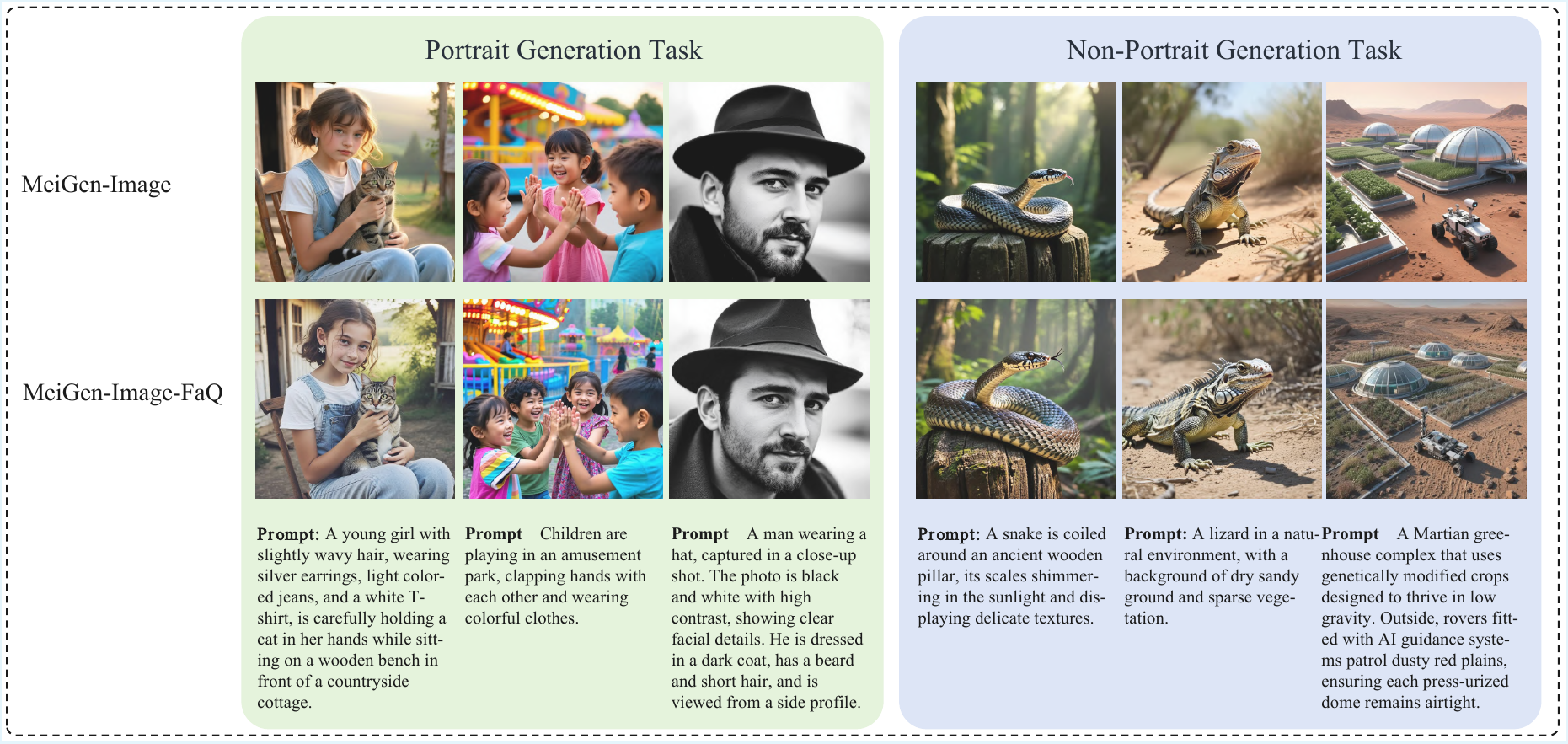} 
\caption{Qualitative cases of MeiGen-Image and MeiGen-Image-FaQ.}
\setlength{\abovecaptionskip}{-5pt}
\label{fig:meigen_case}
\vspace{-10pt}
\end{figure*}

\begin{figure*}[t]
\centering
\includegraphics[width=1.0\textwidth, trim=0cm 0cm 0cm 0cm, clip]{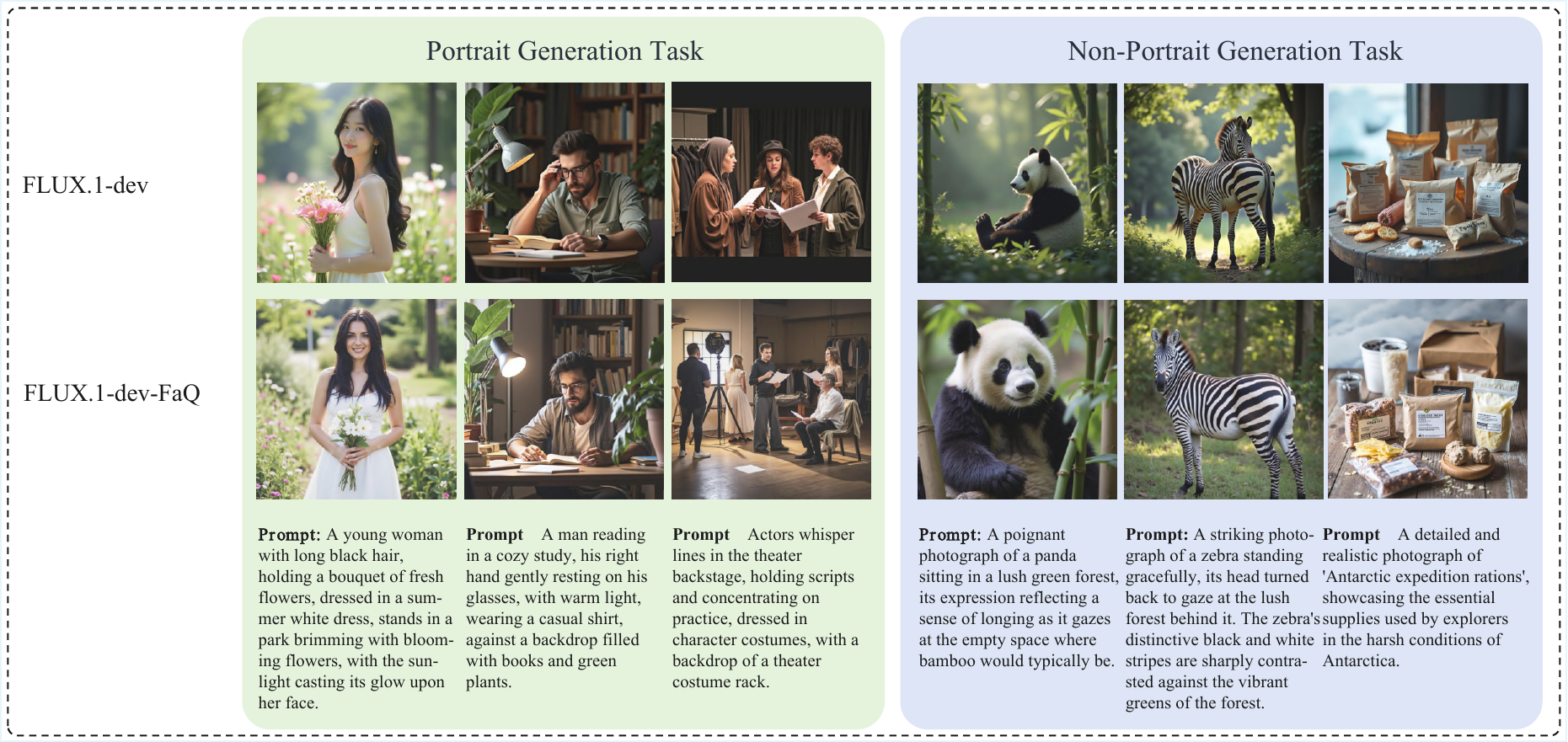} 
\caption{Qualitative cases of FLUX.1-dev and FLUX.1-dev-FaQ.}
\setlength{\abovecaptionskip}{-5pt}
\label{fig:flux_case}
\vspace{-10pt}
\end{figure*}

\subsection{Overall Performance}

\textbf{Performance on benchmarks and human evaluation.}
We evaluate our models on five benchmarks designed to assess three key aspects of generation. For prompt-image alignment, we use GenEval~\citep{ghosh2023geneval} and DPG-Bench~\citep{hu2024ella}. For visual quality, we use COCO-30K FID~\citep{lin2014microsoft} and our custom GPT-Fidelity metric, which employs GPT-4 for pairwise comparisons of image fidelity based on a shared text prompt. For world knowledge reasoning capability, we use WISE~\citep{niu2025wise}.

As presented in Table~\ref{tab:auto_metrics}, Forge-and-Quench significantly boosts visual quality, evidenced by superior scores on COCO-30K FID and GPT-Fidelity across both MeiGen-Image and FLUX.1-dev. Crucially, this enhancement in fidelity comes at no cost to prompt alignment, as our models maintain performance comparable to the original backbones on GenEval and DPG-Bench. This confirms our method's ability to improve the fidelity and detail richness of the image while maintaining robust instruction following. In addition, our models achieve significant improvements on the WISE benchmark, demonstrating its enhancement to the world knowledge reasoning capability.


To complement our automated metrics, we conducted a large-scale human evaluation study across approximately 2,000 prompts. In a side-by-side comparison, annotators were asked to assess image pairs from our model and the original baseline on two criteria: prompt alignment and visual quality. The results, presented in Figure~\ref{fig:human_anno}, are unambiguous: our model achieves performance on par with the original T2I model for prompt alignment, while demonstrating a significant user preference for its superior visual quality.


\textbf{Analysis of visual performance.}
Fig.~\ref{fig:meigen_case} and Fig.~\ref{fig:flux_case} (with additional examples in Appendix~\ref{ap-sec:more_visuals}) showcase qualitative comparisons of our method across different T2I backbones, including MeiGen-Image and FLUX.1-dev. Across both portrait and general scene generation, our framework produces images with markedly improved realism, a significant reduction in common AI artifacts, and superior representation of fine-grained details.

1) MeiGen-Image-FaQ vs. MeiGen-Image: When applied to MeiGen-Image, our framework yields substantial enhancements. Portraits exhibit more realistic skin textures, finer hair details, and more intricate fabric weaves in clothing and accessories. In non-portrait scenes, the generated images show a distinct reduction in artifacts, while high-frequency details in both foreground and background elements are rendered with greater clarity and richness.

2) FLUX.1-dev-FaQ vs. FLUX.1-dev: The benefits of our framework extend to FLUX.1-dev, which also demonstrates enhanced realism, fewer artifacts, and improved detail fidelity. Moreover, our method effectively mitigates several artifacts specific to the FLUX.1-dev model, resulting in a marked reduction in issues such as waxy skin textures, overly stylized cartoon effects, and out-of-focus backgrounds.



\subsection{Ablation Study}


\subsubsection{Bridge Adapter}
\textbf{Architectural design.}
We evaluated three candidate architectures for the Bridge Adapter: Diffusion, AutoRegressive, and direct projection (Fig.\ref{fig:training_strategy}). Table\ref{tab:bridge_type} shows a clear outcome: the diffusion-based approach offers the best trade-off between COCO-30K FID and inference speed. Based on this analysis, we selected the diffusion architecture for our framework.




\begin{table*}[tbp] 
    \begin{minipage}[t]{0.5\linewidth} 
        \centering
        \resizebox{\linewidth}{!}{%
            \begin{tabular}{lcc}
                \toprule 
                Method & FID $\downarrow$ & Latency (s) $\downarrow$ \\
                \midrule 
                MeiGen-Image & 23.97 & - \\
                \addlinespace[3pt] 
                w/ FaQ (Diffusion) & \textbf{19.86} & 0.49 \\
                w/ FaQ (AutoRegressive) & 20.81 & 7.31 \\
                w/ FaQ (Projection) & 21.35 & \textbf{0.10} \\
                \bottomrule
            \end{tabular}
        }
        \renewcommand{\arraystretch}{1.2} 
        \caption{The performance of different Bridge Adapter architectures.}
        \label{tab:bridge_type}
    \end{minipage}%
    \hfill 
    \begin{minipage}[t]{0.45\linewidth} 
        \centering
        \resizebox{\linewidth}{!}{%
            \begin{tabular}{lcc}
                \toprule 
                 & FID $\downarrow$ & Latency (s) $\downarrow$ \\
                \midrule 
                MeiGen-Image (Base) &23.97 & -\\ 
                \addlinespace[3pt]
                \cline{1-3}
                \addlinespace[3pt]
                w/ FaQ (DiT-2B \& QSize=64) & 19.86 & \textbf{0.49} \\
                w/ FaQ (DiT-6B \& QSize=64) & \textbf{19.63} & 1.51 \\
                w/ FaQ (DiT-2B \& QSize=128) & 19.76 & 0.55 \\
                w/ FaQ (DiT-2B \& QSize=256) & 20.08 & 0.62 \\
                \bottomrule
            \end{tabular}%
        }
        \caption{Impact of Bridge Adapter size on model performance.}
        \label{tab:bridge_size}
    \end{minipage}
\vspace{-10pt}
\end{table*}



\textbf{Component size.} We then optimized the size of the diffusion-based adapter by ablating its two key components: Learnable Queries and the DiT module. Our findings in Table~\ref{tab:bridge_size} indicate that performance saturates at a DiT-2B model with a query size of 64. Scaling beyond this point offers negligible gains in quality while increasing computational overhead. This configuration thus represents the optimal point of performance and efficiency.


\begin{figure*}[t]
\centering
\includegraphics[width=1.0\textwidth, trim=0cm 0cm 0cm 0cm, clip]{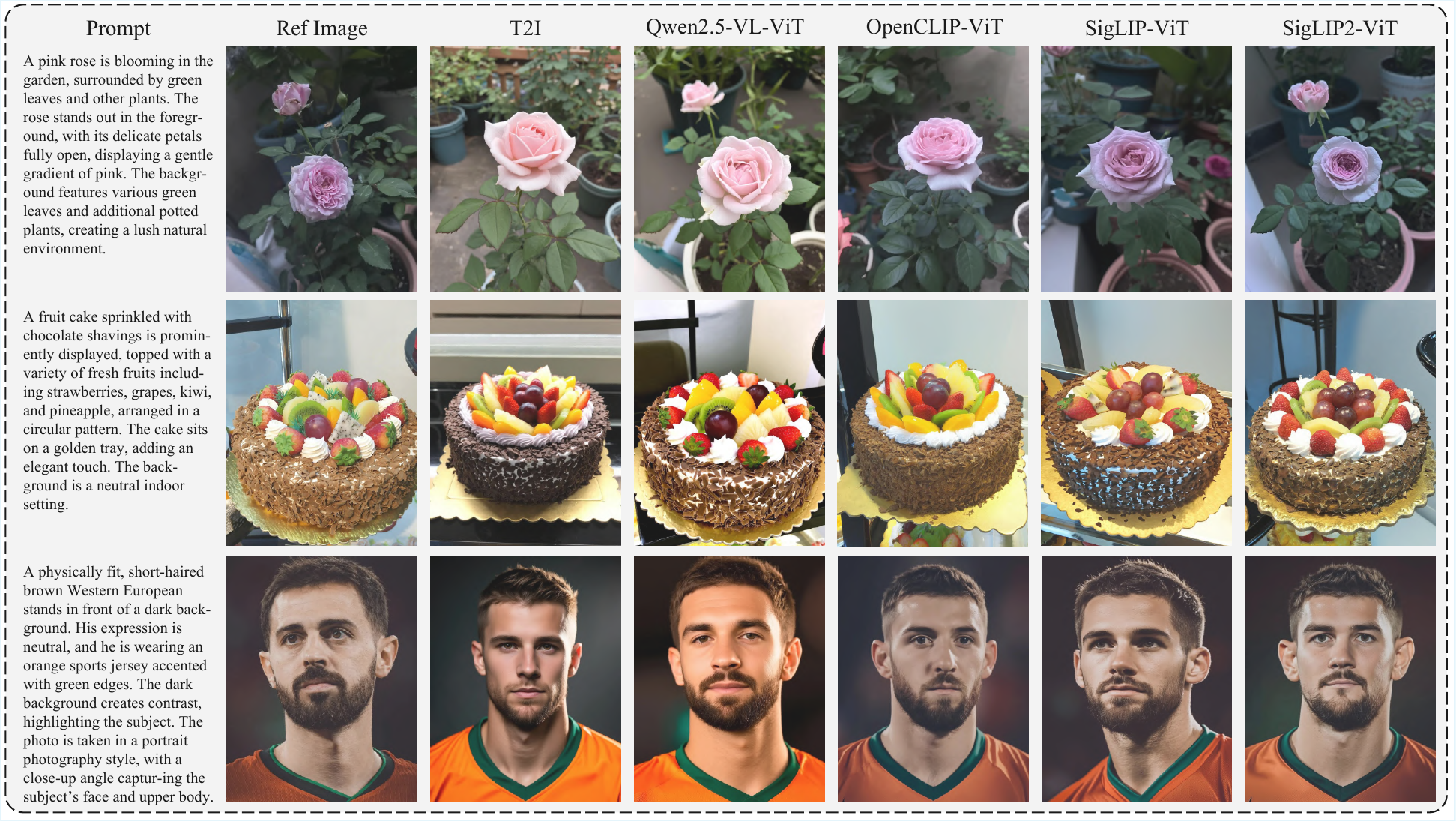} 
\caption{Image visualization based on reference images generated by different visual encoders.}
\label{fig:feature_type}
\vspace{-10pt}
\end{figure*}

\subsubsection{Bridge Feature}
The effectiveness of the Bridge Feature is critically dependent on the choice of the visual encoder that defines its target feature space. In this section, we evaluate four prominent encoders to identify the optimal choice: OpenCLIP-ViT-H-14~\citep{cherti2023reproducible}, Qwen2.5-VL-ViT~\citep{bai2025qwen2}, SigLiP-ViT~\citep{zhai2023sigmoid}, and SigLiP2-ViT~\citep{tschannen2025siglip}.

\textbf{Fidelity enhancement analysis.}
To isolate the intrinsic fidelity-enhancing potential of each visual encoder, we designed an image reconstruction experiment. In this idealized setup, we bypass the Forge stage, instead extracting the bridge feature directly from a ground-truth reference image using each candidate encoder. This feature is then used to condition the \textit{quench} stage.

As shown in Figure~\ref{fig:feature_type}, the results reveal a stark performance gap among the encoders. Features derived from the SigLIP series (SigLIP-ViT and SigLIP2-ViT) enabled reconstructions of significantly higher visual fidelity, closely mirroring the reference images. In stark contrast, features from Qwen2.5-VL-ViT failed to capture meaningful fidelity cues, yielding outputs with artifacts and a quality level indistinguishable from the baseline T2I model. OpenCLIP-ViT's performance was intermediate, offering only marginal fidelity gains.

\begin{figure*}[t]
\centering
\includegraphics[width=1.0\textwidth, trim=0cm 0cm 0cm 0cm, clip]{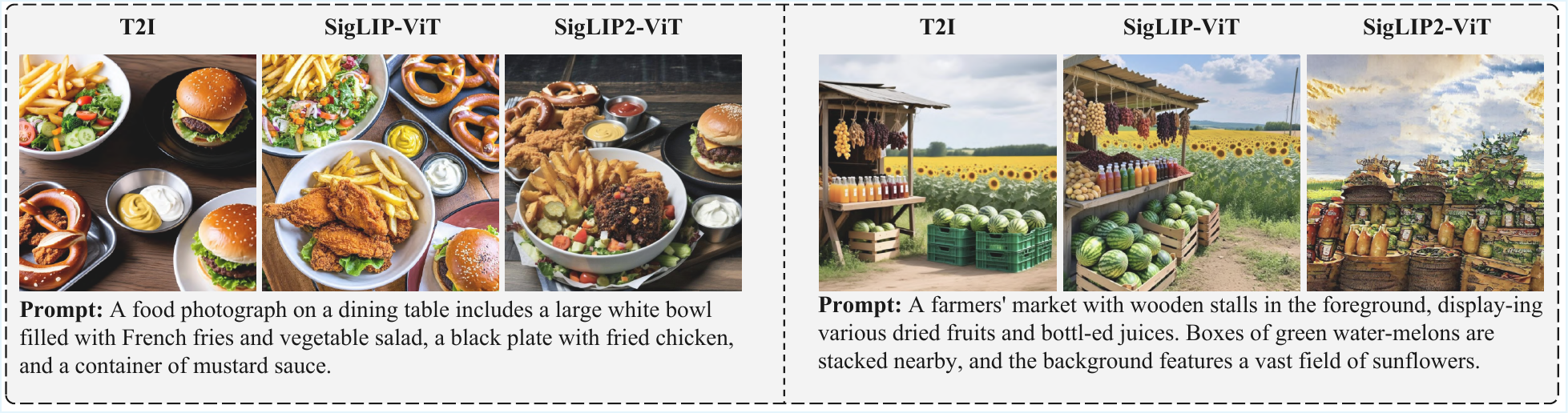} 
\caption{The distortion performance of SigLIP-ViT and SigLIP2-ViT.}
\label{fig:siglip_siglip2}
\end{figure*}

\textbf{Robustness analysis.}
We observed a significant performance gap with SigLIP2-ViT. It excelled in the reference-based reconstruction scenario (Fig.~\ref{fig:feature_type}) where a perfect ground-truth feature is provided. However, in our actual framework which operates on text alone, the forged Bridge feature $e_b$ is not a perfect reconstruction of a real SigLIP feature $e_s$. In this more realistic scenario, SigLIP2-ViT's lack of \textbf{robustness} to the inherent approximation errors becomes apparent, leading to the severe artifacts seen in Fig.~\ref{fig:siglip_siglip2}. We posit that these errors in the forged feature act as noise, to which the SigLIP2-ViT feature space is overly sensitive.


To test this, we perform a noise perturbation analysis, contaminating features with scaled noise drawn from their own statistical distribution:
\begin{equation}
\label{eqn:noised_embs}
\mathbf{e}' = \mathbf{e} + \lambda \cdot \mathcal{N}(\mu_{\mathbf{e}}, \sigma^2_{\mathbf{e}})
\end{equation}
where $\lambda$ is the noise scale. As shown in Table~\ref{tab:cosine_sim}, SigLIP2-ViT's feature similarity degrades significantly faster under noise, confirming its lower robustness. This instability is corroborated by its statistical properties (Table~\ref{tab:feature_values}), which reveal higher variance and sparsity. Such sensitive, brittle features are difficult to forge accurately, leading to visual distortions.

In summary, SigLIP2-ViT's feature instability makes it unsuitable for our framework. We therefore adopt the more stable \textbf{SigLIP-ViT}, which offers the best balance of fidelity and the robustness our two-stage approach requires.

\begin{table*}[tbp] 
    \centering 
    \renewcommand{\arraystretch}{1.2} 

    \begin{minipage}[t]{0.49\textwidth} 
        \centering
        \caption{Cosine similarity between $\mathbf{e}$ and $\mathbf{e'}$ for SigLIP-ViT and SigLIP2-ViT with different $\lambda$.}
        \label{tab:cosine_sim}
        \resizebox{\linewidth}{!}{%
            \begin{tabular}{lcccccc}
                \toprule 
                \textbf{Noise Scale} & \textbf{0.0} & \textbf{0.2} & \textbf{0.4} & \textbf{0.6} & \textbf{0.8} & \textbf{1.0} \\
                \cmidrule(lr){1-7} 
                Cosine Sim. (SigLIP-ViT)  & 1.00 & 0.98 & 0.93 & 0.85 & 0.78 & 0.71 \\ 
                Cosine Sim. (SigLIP2-ViT) & 1.00 & 0.88 & 0.68 & 0.53 & 0.42 & 0.35 \\
                \bottomrule
            \end{tabular}%
        }
    \end{minipage}%
    \hfill 
    \begin{minipage}[t]{0.49\textwidth} 
        \centering
        \caption{Statistical values of features for SigLIP-ViT and SigLIP2-ViT.}
        \label{tab:feature_values}
        \resizebox{\linewidth}{!}{%
            \begin{tabular}{lccccc}
                \toprule 
                 & \textbf{shape} & \textbf{mean} & \textbf{std} & \textbf{norm} & \textbf{abs\_max} \\
                \midrule 
                SigLIP-ViT@384px  & 729×1152 & 0.1060 & 2.0938 & 1920 & 217 \\ 
                SigLIP2-ViT@384px & 576×1152 & 0.0830 & 3.2188 & 2576 & 1680 \\
                \bottomrule
            \end{tabular}%
        }
    \end{minipage}
\end{table*}



\section{Conclusion}
In this work, we presented Forge-and-Quench, a novel framework that significantly boosts the fidelity and detail of images generated by unified multimodal models. Our framework uniquely employs an MLLM to forge two parallel guidance signals: a semantically enriched text prompt and a virtual visual feature that emulates the guidance of a real image embedding.
This dual-conditioning signal is then quenched into a frozen T2I backbone via a lightweight injection adapter, providing fine-grained visual control throughout the generation process. 

Our comprehensive experiments demonstrate that this approach substantially improves image realism and detail without compromising the model's core instruction-following capabilities.
By acting as a universal intermediary, the Bridge Feature enables our lightweight, adapter-based framework to efficiently combine diverse MLLMs and T2I models without retraining, ensuring broad scalability.


\begin{thebibliography}{10}

\bibitem{sun2023emu}
Quan Sun, Qiying Yu, Yufeng Cui, Fan Zhang, Xiaosong Zhang, Yueze Wang, Hongcheng Gao, Jingjing Liu, Tiejun Huang, and Xinlong Wang.
\newblock Emu: Generative pretraining in multimodality.
\newblock {\em arXiv preprint arXiv:2307.05222}, 2023.

\bibitem{sun2024generative}
Quan Sun, Yufeng Cui, Xiaosong Zhang, Fan Zhang, Qiying Yu, Yueze Wang, Yongming Rao, Jingjing Liu, Tiejun Huang, and Xinlong Wang.
\newblock Generative multimodal models are in-context learners.
\newblock In {\em Proceedings of the IEEE/CVF Conference on Computer Vision and Pattern Recognition}, pages 14398--14409, 2024.

\bibitem{team2024chameleon}
Chameleon Team.
\newblock Chameleon: Mixed-modal early-fusion foundation models.
\newblock {\em arXiv preprint arXiv:2405.09818}, 2024.

\bibitem{fan2025unified}
Lijie Fan, Luming Tang, Siyang Qin, Tianhong Li, Xuan Yang, Siyuan Qiao, Andreas Steiner, Chen Sun, Yuanzhen Li, Tao Zhu, et~al.
\newblock Unified autoregressive visual generation and understanding with continuous tokens.
\newblock {\em arXiv preprint arXiv:2503.13436}, 2025.

\bibitem{wu2024vila}
Yecheng Wu, Zhuoyang Zhang, Junyu Chen, Haotian Tang, Dacheng Li, Yunhao Fang, Ligeng Zhu, Enze Xie, Hongxu Yin, Li~Yi, et~al.
\newblock Vila-u: a unified foundation model integrating visual understanding and generation.
\newblock {\em arXiv preprint arXiv:2409.04429}, 2024.

\bibitem{ge2307planting}
Y~Ge, Y~Ge, Z~Zeng, X~Wang, and Y~Shan.
\newblock Planting a seed of vision in large language model. arxiv 2023.
\newblock {\em arXiv preprint arXiv:2307.08041}, 2023.

\bibitem{pan2025transfer}
Xichen Pan, Satya~Narayan Shukla, Aashu Singh, Zhuokai Zhao, Shlok~Kumar Mishra, Jialiang Wang, Zhiyang Xu, Jiuhai Chen, Kunpeng Li, Felix Juefei-Xu, et~al.
\newblock Transfer between modalities with metaqueries.
\newblock {\em arXiv preprint arXiv:2504.06256}, 2025.

\bibitem{chen2025blip3}
Jiuhai Chen, Zhiyang Xu, Xichen Pan, Yushi Hu, Can Qin, Tom Goldstein, Lifu Huang, Tianyi Zhou, Saining Xie, Silvio Savarese, et~al.
\newblock Blip3-o: A family of fully open unified multimodal models-architecture, training and dataset.
\newblock {\em arXiv preprint arXiv:2505.09568}, 2025.

\bibitem{zhang2023adding}
Lvmin Zhang, Anyi Rao, and Maneesh Agrawala.
\newblock Adding conditional control to text-to-image diffusion models.
\newblock In {\em Proceedings of the IEEE/CVF international conference on computer vision}, pages 3836--3847, 2023.

\bibitem{ye2023ip}
Hu~Ye, Jun Zhang, Sibo Liu, Xiao Han, and Wei Yang.
\newblock Ip-adapter: Text compatible image prompt adapter for text-to-image diffusion models.
\newblock {\em arXiv preprint arXiv:2308.06721}, 2023.

\bibitem{lian2023llmgrounded}
Long Lian, Boyi Li, Adam Yala, and Trevor Darrell.
\newblock Llm-grounded diffusion: Enhancing prompt understanding of text-to-image diffusion models with large language models.
\newblock {\em arXiv preprint arXiv:2305.13655}, 2023.

\bibitem{wu2024next}
Shengqiong Wu, Hao Fei, Leigang Qu, Wei Ji, and Tat-Seng Chua.
\newblock Next-gpt: Any-to-any multimodal llm.
\newblock In {\em Forty-first International Conference on Machine Learning}, 2024.

\bibitem{zhou2024transfusion}
Chunting Zhou, Lili Yu, Arun Babu, Kushal Tirumala, Michihiro Yasunaga, Leonid Shamis, Jacob Kahn, Xuezhe Ma, Luke Zettlemoyer, and Omer Levy.
\newblock Transfusion: Predict the next token and diffuse images with one multi-modal model.
\newblock {\em arXiv preprint arXiv:2408.11039}, 2024.

\bibitem{liao2025mogao}
Chao Liao, Liyang Liu, Xun Wang, Zhengxiong Luo, Xinyu Zhang, Wenliang Zhao, Jie Wu, Liang Li, Zhi Tian, and Weilin Huang.
\newblock Mogao: An omni foundation model for interleaved multi-modal generation.
\newblock {\em arXiv preprint arXiv:2505.05472}, 2025.

\bibitem{deng2025emerging}
Chaorui Deng, Deyao Zhu, Kunchang Li, Chenhui Gou, Feng Li, Zeyu Wang, Shu Zhong, Weihao Yu, Xiaonan Nie, Ziang Song, et~al.
\newblock Emerging properties in unified multimodal pretraining.
\newblock {\em arXiv preprint arXiv:2505.14683}, 2025.

\bibitem{wu2025janus}
Chengyue Wu, Xiaokang Chen, Zhiyu Wu, Yiyang Ma, Xingchao Liu, Zizheng Pan, Wen Liu, Zhenda Xie, Xingkai Yu, Chong Ruan, et~al.
\newblock Janus: Decoupling visual encoding for unified multimodal understanding and generation.
\newblock In {\em Proceedings of the Computer Vision and Pattern Recognition Conference}, pages 12966--12977, 2025.

\bibitem{chen2025janus}
Xiaokang Chen, Zhiyu Wu, Xingchao Liu, Zizheng Pan, Wen Liu, Zhenda Xie, Xingkai Yu, and Chong Ruan.
\newblock Janus-pro: Unified multimodal understanding and generation with data and model scaling.
\newblock {\em arXiv preprint arXiv:2501.17811}, 2025.

\bibitem{ge2023making}
Yuying Ge, Sijie Zhao, Ziyun Zeng, Yixiao Ge, Chen Li, Xintao Wang, and Ying Shan.
\newblock Making llama see and draw with seed tokenizer.
\newblock {\em arXiv preprint arXiv:2310.01218}, 2023.

\bibitem{ge2024seed}
Yuying Ge, Sijie Zhao, Jinguo Zhu, Yixiao Ge, Kun Yi, Lin Song, Chen Li, Xiaohan Ding, and Ying Shan.
\newblock Seed-x: Multimodal models with unified multi-granularity comprehension and generation.
\newblock {\em arXiv preprint arXiv:2404.14396}, 2024.

\bibitem{nichol2021glide}
Alex Nichol, Prafulla Dhariwal, Aditya Ramesh, Pranav Shyam, Pamela Mishkin, Bob McGrew, Ilya Sutskever, and Mark Chen.
\newblock Glide: Towards photorealistic image generation and editing with text-guided diffusion models.
\newblock {\em arXiv preprint arXiv:2112.10741}, 2021.

\bibitem{ramesh2022hierarchical}
Aditya Ramesh, Prafulla Dhariwal, Alex Nichol, Casey Chu, and Mark Chen.
\newblock Hierarchical text-conditional image generation with clip latents.
\newblock {\em arXiv preprint arXiv:2204.06125}, 1(2):3, 2022.

\bibitem{rombach2022high}
Robin Rombach, Andreas Blattmann, Dominik Lorenz, Patrick Esser, and Bj{\"o}rn Ommer.
\newblock High-resolution image synthesis with latent diffusion models.
\newblock In {\em Proceedings of the IEEE/CVF conference on computer vision and pattern recognition}, pages 10684--10695, 2022.

\bibitem{saharia2022photorealistic}
Chitwan Saharia, William Chan, Saurabh Saxena, Lala Li, Jay Whang, Emily~L Denton, Kamyar Ghasemipour, Raphael Gontijo~Lopes, Burcu Karagol~Ayan, Tim Salimans, et~al.
\newblock Photorealistic text-to-image diffusion models with deep language understanding.
\newblock {\em Advances in neural information processing systems}, 35:36479--36494, 2022.

\bibitem{podell2023sdxl}
Dustin Podell, Zion English, Kyle Lacey, Andreas Blattmann, Tim Dockhorn, Jonas M{\"u}ller, Joe Penna, and Robin Rombach.
\newblock Sdxl: Improving latent diffusion models for high-resolution image synthesis.
\newblock {\em arXiv preprint arXiv:2307.01952}, 2023.

\bibitem{kolors}
Kolors Team.
\newblock Kolors: Effective training of diffusion model for photorealistic text-to-image synthesis.
\newblock {\em arXiv preprint}, 2024.

\bibitem{midjourney-v6.1}
midjourney team.
\newblock midjourney-v6.1 official website, 2024.

\bibitem{cai2025hidream}
Qi~Cai, Jingwen Chen, Yang Chen, Yehao Li, Fuchen Long, Yingwei Pan, Zhaofan Qiu, Yiheng Zhang, Fengbin Gao, Peihan Xu, et~al.
\newblock Hidream-i1: A high-efficient image generative foundation model with sparse diffusion transformer.
\newblock {\em arXiv preprint arXiv:2505.22705}, 2025.

\bibitem{esser2024scaling}
Patrick Esser, Sumith Kulal, Andreas Blattmann, Rahim Entezari, Jonas M{\"u}ller, Harry Saini, Yam Levi, Dominik Lorenz, Axel Sauer, Frederic Boesel, et~al.
\newblock Scaling rectified flow transformers for high-resolution image synthesis.
\newblock In {\em Forty-first international conference on machine learning}, 2024.

\bibitem{blackforestlabs_flux}
Black~Forest Labs.
\newblock Flux: Official inference repository for flux.1 models, 2024.

\bibitem{gao2025seedream}
Yu~Gao, Lixue Gong, Qiushan Guo, Xiaoxia Hou, Zhichao Lai, Fanshi Li, Liang Li, Xiaochen Lian, Chao Liao, Liyang Liu, et~al.
\newblock Seedream 3.0 technical report.
\newblock {\em arXiv preprint arXiv:2504.11346}, 2025.

\bibitem{wu2025qwen}
Chenfei Wu, Jiahao Li, Jingren Zhou, Junyang Lin, Kaiyuan Gao, Kun Yan, Sheng-ming Yin, Shuai Bai, Xiao Xu, Yilei Chen, et~al.
\newblock Qwen-image technical report.
\newblock {\em arXiv preprint arXiv:2508.02324}, 2025.

\bibitem{wang2025image}
Jia Wang, Jie Hu, Xiaoqi Ma, Hanghang Ma, Xiaoming Wei, and Enhua Wu.
\newblock Image editing with diffusion models: A survey.
\newblock {\em arXiv preprint arXiv:2504.13226}, 2025.

\bibitem{mou2024t2i}
Chong Mou, Xintao Wang, Liangbin Xie, Yanze Wu, Jian Zhang, Zhongang Qi, and Ying Shan.
\newblock T2i-adapter: Learning adapters to dig out more controllable ability for text-to-image diffusion models.
\newblock In {\em Proceedings of the AAAI conference on artificial intelligence}, volume~38, pages 4296--4304, 2024.

\bibitem{huang2023composer}
Lianghua Huang, Di~Chen, Yu~Liu, Yujun Shen, Deli Zhao, and Jingren Zhou.
\newblock Composer: Creative and controllable image synthesis with composable conditions.
\newblock {\em arXiv preprint arXiv:2302.09778}, 2023.

\bibitem{fu2023guiding}
Tsu-Jui Fu, Wenze Hu, Xianzhi Du, William~Yang Wang, Yinfei Yang, and Zhe Gan.
\newblock Guiding instruction-based image editing via multimodal large language models.
\newblock {\em arXiv preprint arXiv:2309.17102}, 2023.

\bibitem{he2024freeedit}
Runze He, Kai Ma, Linjiang Huang, Shaofei Huang, Jialin Gao, Xiaoming Wei, Jiao Dai, Jizhong Han, and Si~Liu.
\newblock Freeedit: Mask-free reference-based image editing with multi-modal instruction.
\newblock {\em arXiv preprint arXiv:2409.18071}, 2024.

\bibitem{lipman2022flow}
Yaron Lipman, Ricky~TQ Chen, Heli Ben-Hamu, Maximilian Nickel, and Matt Le.
\newblock Flow matching for generative modeling.
\newblock {\em arXiv preprint arXiv:2210.02747}, 2022.

\bibitem{liu2022flow}
Xingchao Liu, Chengyue Gong, and Qiang Liu.
\newblock Flow straight and fast: Learning to generate and transfer data with rectified flow.
\newblock {\em arXiv preprint arXiv:2209.03003}, 2022.

\bibitem{zhai2023sigmoid}
Xiaohua Zhai, Basil Mustafa, Alexander Kolesnikov, and Lucas Beyer.
\newblock Sigmoid loss for language image pre-training.
\newblock In {\em Proceedings of the IEEE/CVF international conference on computer vision}, pages 11975--11986, 2023.

\bibitem{ghosh2023geneval}
Dhruba Ghosh, Hannaneh Hajishirzi, and Ludwig Schmidt.
\newblock Geneval: An object-focused framework for evaluating text-to-image alignment.
\newblock {\em Advances in Neural Information Processing Systems}, 2023.

\bibitem{hu2024ella}
Xiwei Hu, Rui Wang, Yixiao Fang, Bin Fu, Pei Cheng, and Gang Yu.
\newblock Ella: Equip diffusion models with llm for enhanced semantic alignment.
\newblock {\em arXiv preprint arXiv:2403.05135}, 2024.

\bibitem{lin2014microsoft}
Tsung-Yi Lin, Michael Maire, Serge Belongie, James Hays, Pietro Perona, Deva Ramanan, Piotr Doll{\'a}r, and C~Lawrence Zitnick.
\newblock Microsoft coco: Common objects in context.
\newblock In {\em European conference on computer vision}, pages 740--755. Springer, 2014.

\bibitem{niu2025wise}
Yuwei Niu, Munan Ning, Mengren Zheng, Weiyang Jin, Bin Lin, Peng Jin, Jiaqi Liao, Chaoran Feng, Kunpeng Ning, Bin Zhu, et~al.
\newblock Wise: A world knowledge-informed semantic evaluation for text-to-image generation.
\newblock {\em arXiv preprint arXiv:2503.07265}, 2025.

\bibitem{cherti2023reproducible}
Mehdi Cherti, Romain Beaumont, Ross Wightman, Mitchell Wortsman, Gabriel Ilharco, Cade Gordon, Christoph Schuhmann, Ludwig Schmidt, and Jenia Jitsev.
\newblock Reproducible scaling laws for contrastive language-image learning.
\newblock In {\em Proceedings of the IEEE/CVF conference on computer vision and pattern recognition}, pages 2818--2829, 2023.

\bibitem{bai2025qwen2}
Shuai Bai, Keqin Chen, Xuejing Liu, Jialin Wang, Wenbin Ge, Sibo Song, Kai Dang, Peng Wang, Shijie Wang, Jun Tang, et~al.
\newblock Qwen2.5-vl technical report.
\newblock {\em arXiv preprint arXiv:2502.13923}, 2025.

\bibitem{tschannen2025siglip}
Michael Tschannen, Alexey Gritsenko, Xiao Wang, Muhammad~Ferjad Naeem, Ibrahim Alabdulmohsin, Nikhil Parthasarathy, Talfan Evans, Lucas Beyer, Ye~Xia, Basil Mustafa, et~al.
\newblock Siglip 2: Multilingual vision-language encoders with improved semantic understanding, localization, and dense features.
\newblock {\em arXiv preprint arXiv:2502.14786}, 2025.

\end{thebibliography}

\newpage

\appendix
\section{Appendix}
This supplementary material is structured into several sections to provide additional details and analysis for our work. Specifically, it covers the following topics:
\begin{itemize}
\item In Appendix~\ref{ap-sec:training_setting}, we provide the detailed hyperparameters for training both the Forge and Quench parts of our framework.
\item In Appendix~\ref{ap-sec:demo_design}, we briefly outline the design of our demonstration system.
\item In Appendix~\ref{ap-sec:detailed_results}, we present comprehensive and detailed results on the GenEval, DPG-Bench, and WISE benchmarks.
\item In Appendix~\ref{ap-sec:more_visuals}, we showcase additional qualitative examples to further demonstrate the performance improvements of our method on the MeiGen-Image and FLUX.1-dev.
\end{itemize}

\subsection{Training Setting}
\label{ap-sec:training_setting}
The detailed hyperparameters for training the Forge and Quench components are summarized in Table \ref{tab:hyperparameters}. The Forge part, which includes the Bridge Adapter, was trained on a larger dataset to learn the mapping from MLLM embeddings to the visual feature space. The Quench part, comprising the Injection Adapter, was trained on a filtered, smaller dataset to adapt the T2I model for the new visual condition.

\begin{table}[h]
    \renewcommand{\arraystretch}{1.2}
    \caption{Detailed hyperparameters for training.}
    \label{tab:hyperparameters}
    \centering
    \begin{tabular}{lcc}
        \toprule 
             &Forge & Quench \\
        \midrule 
            Module Size &2B &1B\\ 
            Training Data & 200M & 13M \\
            LR & 1e-4 & 1e-4 \\
            LR Scheduler & Constant & Constant \\
            Training Steps & 500k & 80k (50k@512px + 30k@1024px) \\
            Global Batchsize & 512 & 256 \\
            Optimizer & AdamW(beta1=0.9, beta2=0.95)  & AdamW(beta1=0.9, beta2=0.95)\\
        \bottomrule
    \end{tabular}
\end{table}

\subsection{Demo Design}
\label{ap-sec:demo_design}
To provide a tangible and intuitive illustration of the Forge-and-Quench framework's advantages, we have developed an interactive demonstration system. This platform allows users to input their own custom text prompts and receive immediate, real-time visual feedback. Crucially, the interface presents a direct, side-by-side comparison, simultaneously displaying the image generated by a baseline T2I model alongside the output from our enhanced model. This comparative layout is specifically designed to highlight and validate the significant improvements our framework delivers in terms of image fidelity, the rendering of fine-grained details, and overall prompt alignment.

\begin{figure*}[t]
\centering
\includegraphics[width=0.85\textwidth, trim=0cm 0cm 0cm 0cm, clip]{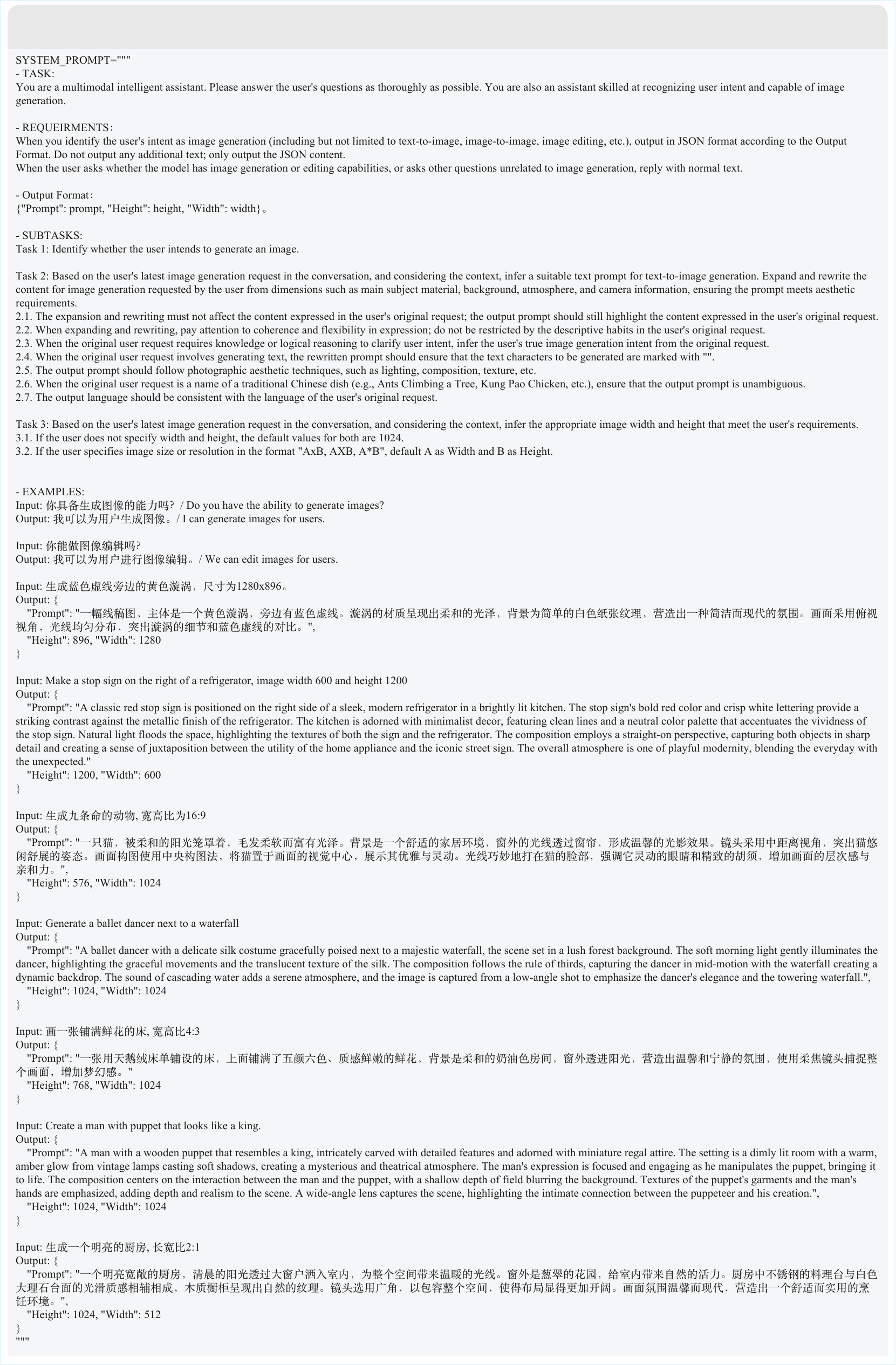} 
\caption{The system prompt of our chat demo.}
\label{fig:sys_prompt}
\end{figure*}

\begin{figure*}[t]
\centering
\includegraphics[width=0.9\textwidth, trim=0cm 0cm 0cm 0cm, clip]{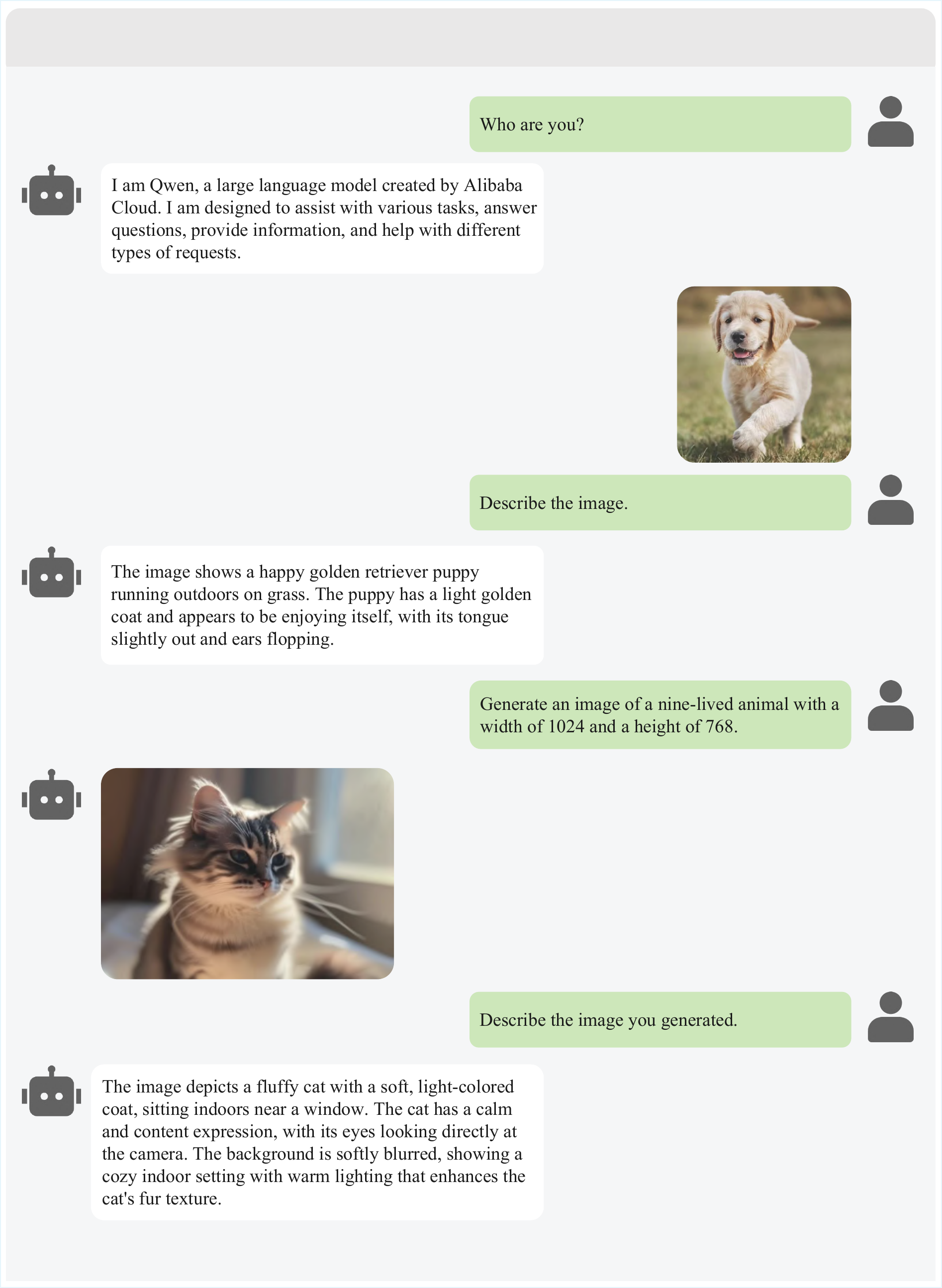} 
\caption{The interactive interface of the chat demo.}
\label{fig:chat_demo}
\end{figure*}

\begin{table}[h]
    \renewcommand{\arraystretch}{1.2}
    \caption{Detailed results on GenEval benchmark.}
    \label{tab:detailed_geneval}
    \centering
    \resizebox{\textwidth}{!}{ 
    \begin{tabular}{lccccccc}
        \toprule 
            \textbf{Method}& \textbf{Single Object}& \textbf{Two Object}& \textbf{Counting}& \textbf{Colors}& \textbf{Position}& \textbf{Color Attr}& \textbf{Overall}\\
        \midrule 
            MeiGen-Image &0.9906 &0.9217 &0.7375 &0.9069 &0.4750 &0.6750 &0.7845\\ 
            MeiGen-Image-FaQ &0.9875 &0.9520 &0.7531 &0.8989 &0.4934 &0.6175 &0.7837\\
        \addlinespace[3pt]
        \cline{1-8}
        \addlinespace[3pt]
            FLUX.1-dev &0.9844 &0.7955 &0.6531 &0.7952 &0.2375 &0.4450 &0.6518\\ 
            FLUX.1-dev-FaQ &1.0000 &0.7831 &0.6366 &0.7897 &0.2145 &0.4377 &0.6436\\ 
        \bottomrule
    \end{tabular}
    }
\end{table}

\begin{table}[h]
    \renewcommand{\arraystretch}{1.2}
    \caption{Detailed results on DPG-Bench.}
    \label{tab:detailed_dpg}
    \centering
    \resizebox{0.75\textwidth}{!}{ 
    \begin{tabular}{lcccccc}
        \toprule 
            \textbf{Method}& \textbf{Global}& \textbf{Entity}& \textbf{Attribute}& \textbf{Relation}& \textbf{Other}& \textbf{Overall}\\
        \midrule 
            MeiGen-Image &85.11 &91.73 &88.80 &93.36 &83.60 &85.94\\
            MeiGen-Image-FaQ &85.41 &92.12 &89.18 &93.89 &87.60 &86.83\\
        \addlinespace[3pt]
        \cline{1-7}
        \addlinespace[3pt]
            FLUX.1-dev &82.67 &89.81 &86.97 &92.80 &82.00 &83.66\\
            FLUX.1-dev-FaQ &81.46 &89.67 &87.05 &93.04 &83.20 &83.01\\
        \bottomrule
    \end{tabular}
    }
\end{table}

\begin{table}[h]
    \renewcommand{\arraystretch}{1.2}
    \caption{Detailed results on WISE benchmark.}
    \label{tab:detailed_wise}
    \centering
    \resizebox{0.85\textwidth}{!}{ 
    \begin{tabular}{lccccccc}
        \toprule 
            \textbf{Method}& \textbf{Cultural}& \textbf{Time}& \textbf{Space}& \textbf{Biology}& \textbf{Physics}& \textbf{Chemistry}& \textbf{Overall}\\
        \midrule 
            MeiGen-Image &0.54 &0.57 &0.66 &0.48 &0.60 &0.40 &0.55\\ 
            MeiGen-Image-FaQ &0.74 &0.66 &0.77 &0.64 &0.72 &0.57 &0.70\\
        \addlinespace[3pt]
        \cline{1-8}
        \addlinespace[3pt]
            FLUX.1-dev &0.55 &0.60 &0.69 &0.45 &0.58 &0.41 &0.56\\ 
            FLUX.1-dev-FaQ &0.70 &0.65 &0.70 &0.62 &0.70 &0.51 &0.66\\ 
        \bottomrule
    \end{tabular}
    }
\end{table}

\subsection{Detailed Results on Benchmarks}
\label{ap-sec:detailed_results}
To provide a more granular view of our model's performance, this section presents detailed breakdowns of the results on several key benchmarks. Table~\ref{tab:detailed_geneval} shows the performance across different categories of the GenEval benchmark. Table~\ref{tab:detailed_dpg} provides a detailed analysis from the DPG-Bench. Finally, Table~\ref{tab:detailed_wise} breaks down the scores on the WISE benchmark, evaluating performance across various domains such as culture, science, and biology.

\subsection{More Visual Results}
\label{ap-sec:more_visuals}

\begin{figure*}[t]
\centering
\includegraphics[width=0.85\textwidth, trim=0cm 0cm 0cm 0cm, clip]{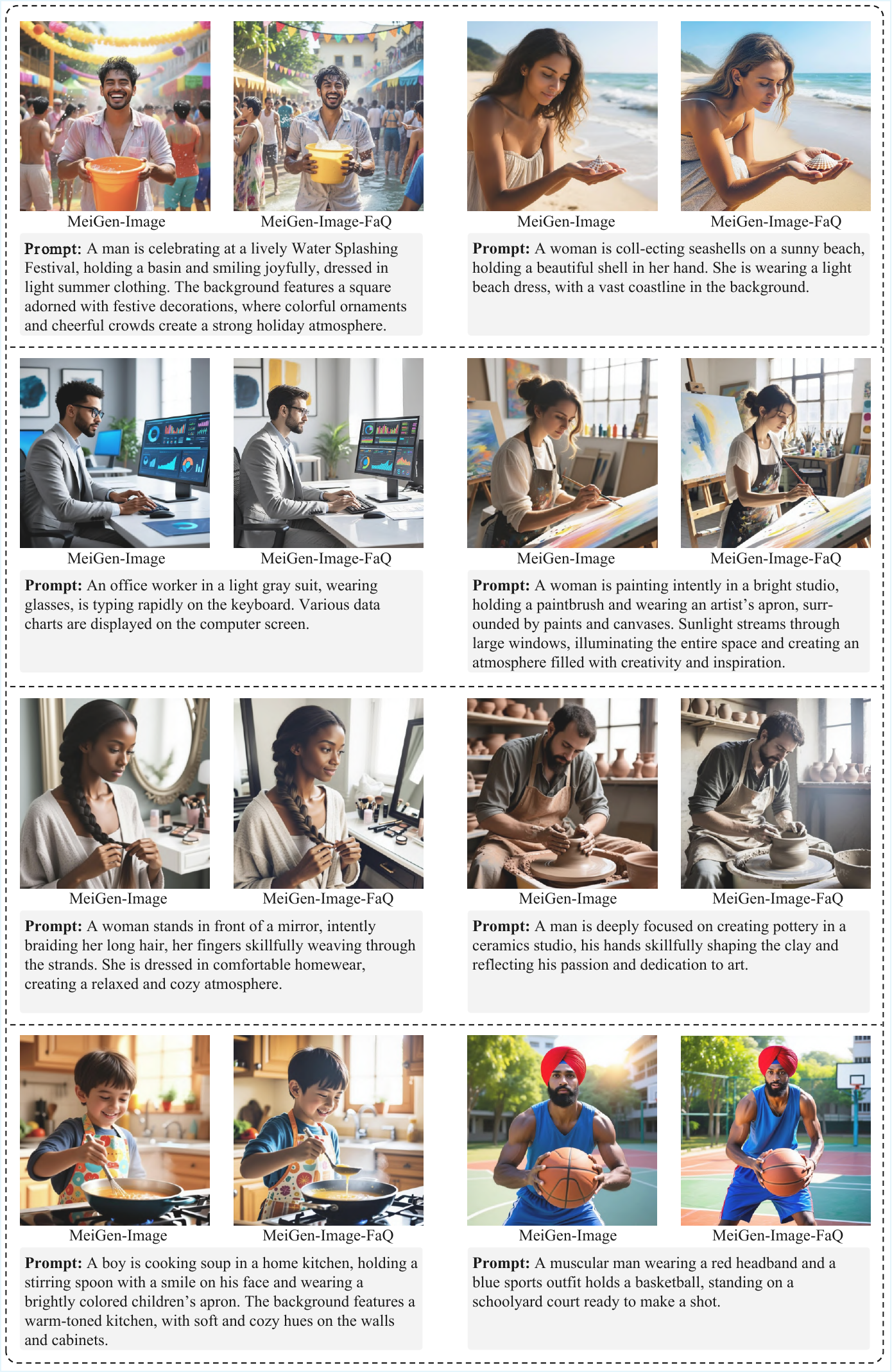} 
\caption{Qualitative cases of MeiGen-Image and MeiGen-Image-FaQ (P1).}
\label{fig:meigen_case_p1}
\end{figure*}

\begin{figure*}[t]
\centering
\includegraphics[width=0.85\textwidth, trim=0cm 0cm 0cm 0cm, clip]{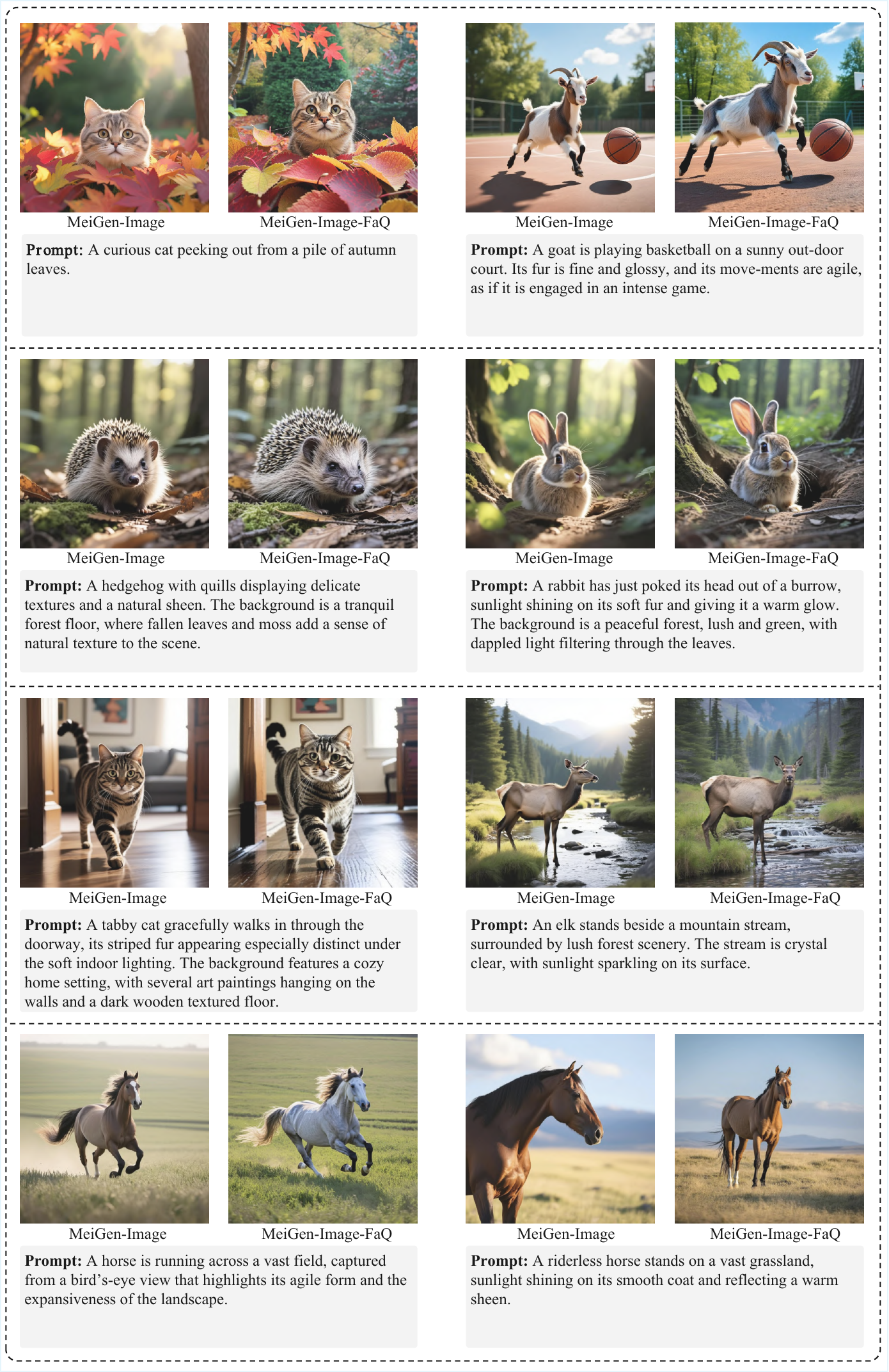} 
\caption{Qualitative cases of MeiGen-Image and MeiGen-Image-FaQ (P2).}
\label{fig:meigen_case_p2}
\end{figure*}

\begin{figure*}[t]
\centering
\includegraphics[width=0.85\textwidth, trim=0cm 0cm 0cm 0cm, clip]{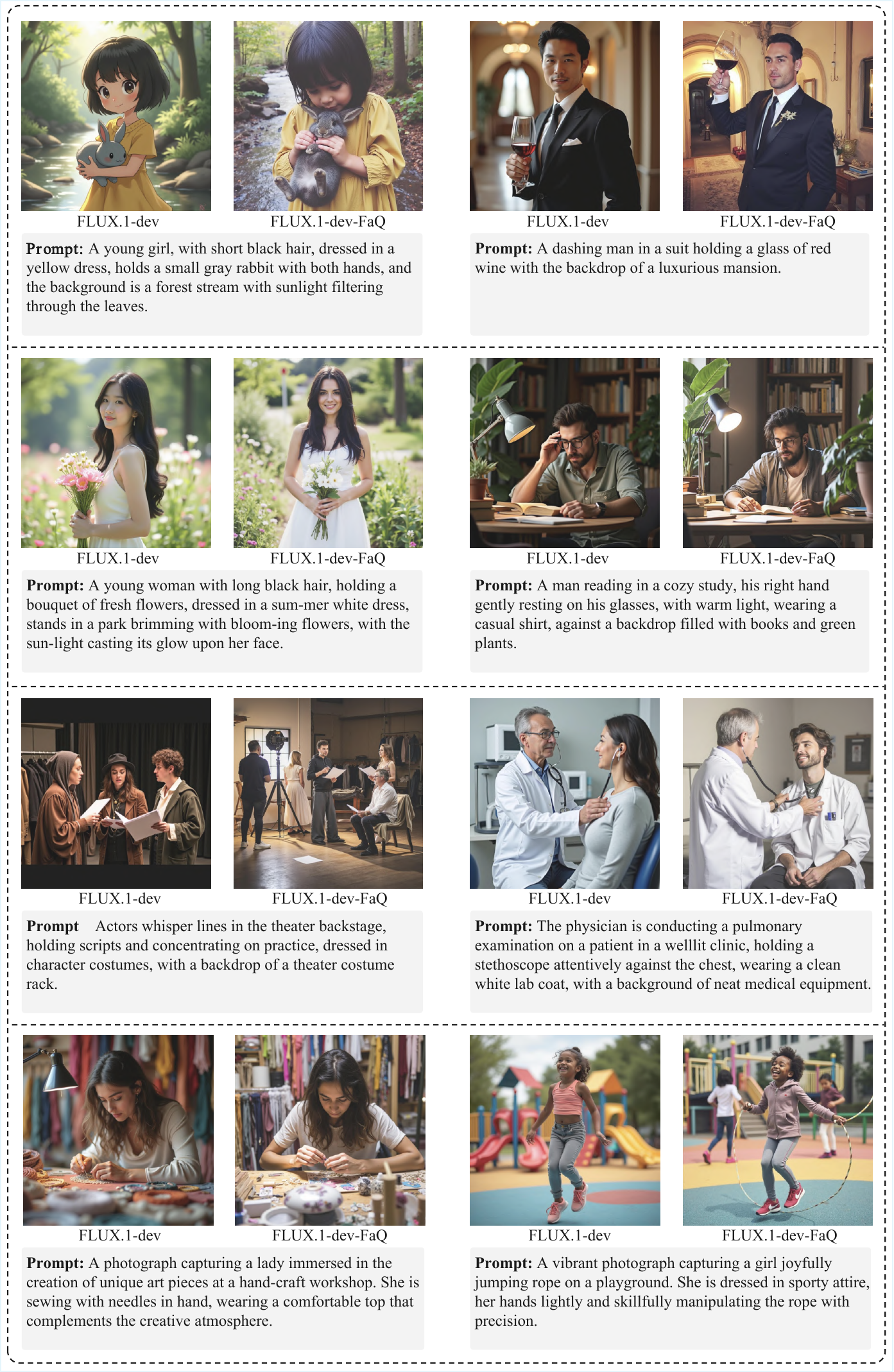} 
\caption{Qualitative cases of FLUX.1-dev and FLUX.1-dev-FaQ (P1).}
\label{fig:flux_case_p1}
\end{figure*}

\begin{figure*}[t]
\centering
\includegraphics[width=0.85\textwidth, trim=0cm 0cm 0cm 0cm, clip]{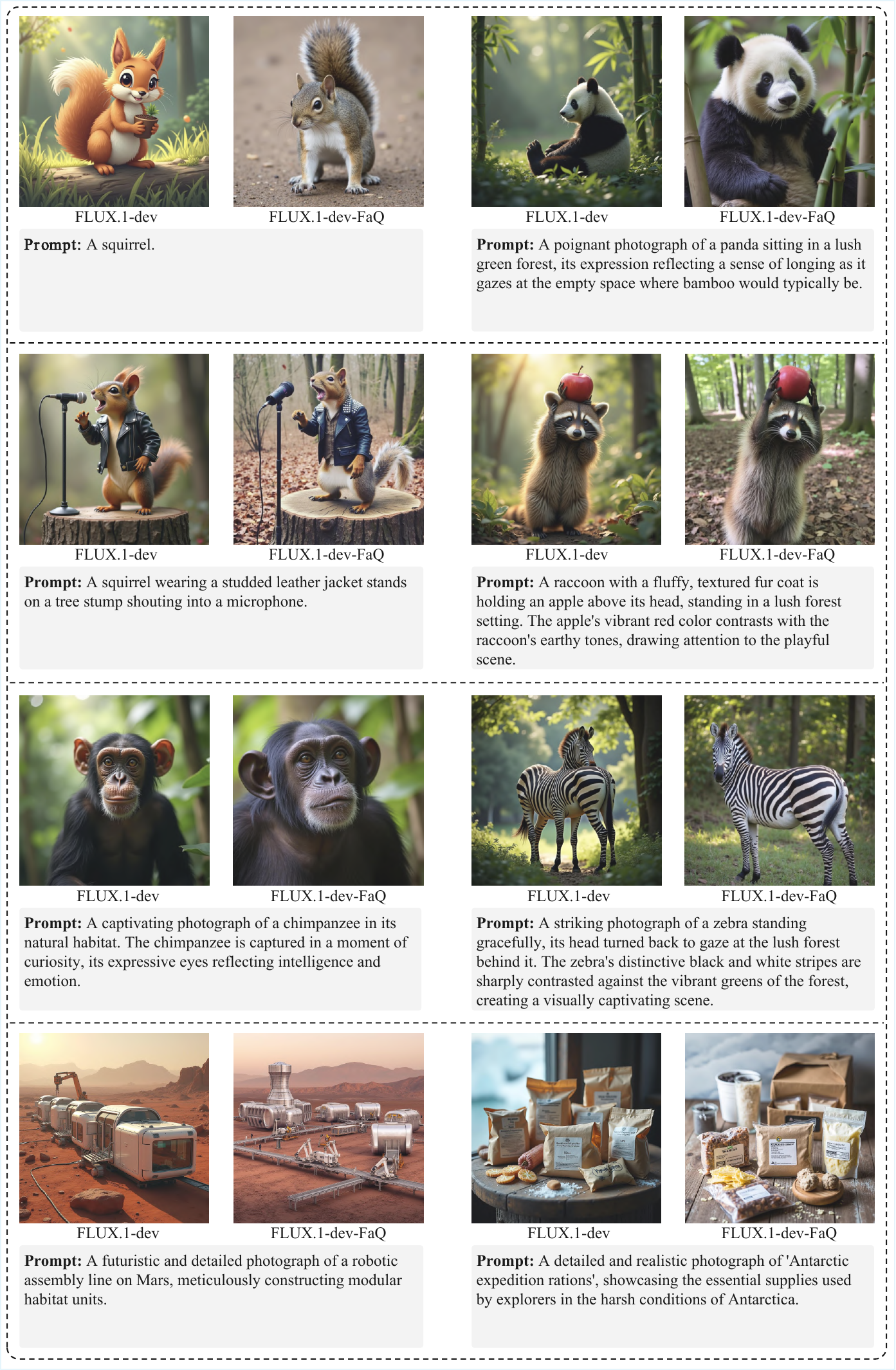} 
\caption{Qualitative cases of FLUX.1-dev and FLUX.1-dev-FaQ (P2).}
\label{fig:flux_case_p2}
\end{figure*}

To visually supplement our quantitative findings, Fig.~\ref{fig:meigen_case_p1} and Fig.~\ref{fig:meigen_case_p2} present more side-by-side comparisons of images generated by the original MeiGen-Image and our enhanced MeiGen-Image-FaQ model, and Fig.~\ref{fig:flux_case_p1} and Fig.~\ref{fig:flux_case_p2} present that of the original FLUX.1-dev and our enhanced FLUX.1-dev-FaQ model. These examples cover a diverse range of prompts, including both portrait and non-portrait scenes, demonstrating consistent improvements in realism, texture detail, and overall aesthetic quality.



\end{document}